\documentclass[conference]{IEEEtran}
\IEEEoverridecommandlockouts
\usepackage{cite}
\usepackage{amsmath,amssymb,amsfonts}

\usepackage{graphicx}
\usepackage{textcomp}
\usepackage{booktabs, multirow, multicol}

\usepackage{algorithm}
 \usepackage{algpseudocode}

\usepackage{amsthm}
\theoremstyle{definition}
\newtheorem{definition}{Definition}[section]
\usepackage{xcolor}
\def\BibTeX{{\rm B\kern-.05em{\sc i\kern-.025em b}\kern-.08em
    T\kern-.1667em\lower.7ex\hbox{E}\kern-.125emX}}

\makeatletter

\let\old@ps@IEEEtitlepagestyle\ps@IEEEtitlepagestyle
\def\confheader#1{%
    \def\ps@IEEEtitlepagestyle{%
        \old@ps@IEEEtitlepagestyle%
        \def\@oddhead{\strut\hfill#1\hfill\strut}%
        \def\@evenhead{\strut\hfill#1\hfill\strut}%
    }%
    \ps@headings%
}
\makeatother

\confheader{%
    \parbox{20cm}{Accepted by 2024 IEEE International Conference on Communications (ICC), \textcopyright 2023 IEEE}
}
\begin{document}

\title{AI in Energy Digital Twining: A Reinforcement Learning-based Adaptive Digital Twin Model for Green Cities \\
}
\author{
	\IEEEauthorblockN{Lal Verda Cakir\IEEEauthorrefmark{1}\IEEEauthorrefmark{2}, Kubra Duran\IEEEauthorrefmark{1}\IEEEauthorrefmark{3}, Craig Thomson\IEEEauthorrefmark{1}, Matthew Broadbent\IEEEauthorrefmark{1}, and Berk Canberk\IEEEauthorrefmark{1}}
	\IEEEauthorblockA{\IEEEauthorrefmark{1}School of Computing, Engineering and The Built Environment, Edinburgh Napier University, UK
	}
 \IEEEauthorblockA{\IEEEauthorrefmark{2} Department of Computer Engineering, Istanbul Technical University, Istanbul, Turkey
	}
  \IEEEauthorblockA{\IEEEauthorrefmark{3}BTS Group, Istanbul, Turkey
	}
    
\{lal.cakir, kubra.duran, C.Thomson3, M.Broadbent, B.Canberk\}@napier.ac.uk, \\ cakirl18@itu.edu.tr, kubra.duran@btsgrp.com

}

\maketitle

\begin{abstract}
Digital Twins (DT) have become crucial to achieve sustainable and effective smart urban solutions. However, current DT modelling techniques cannot support the dynamicity of these smart city environments. This is caused by the lack of right-time data capturing in traditional approaches, resulting in inaccurate modelling and high resource and energy consumption challenges. To fill this gap, we explore spatiotemporal graphs and propose the Reinforcement Learning-based Adaptive Twining (RL-AT) mechanism with Deep Q Networks (DQN). By doing so, our study contributes to advancing Green Cities and showcases tangible benefits in accuracy, synchronisation, resource optimization, and energy efficiency. As a result, we note the spatiotemporal graphs are able to offer a consistent accuracy and 55\% higher querying performance when implemented using graph databases. In addition, our model demonstrates right-time data capturing with 20\% lower overhead and 25\% lower energy consumption. 
\end{abstract}


\section{Introduction}


The pressing challenge of climate change, rapid population growth, and resource scarcity necessitates green smart city solutions that can integrate sustainable and efficient solutions \cite{smart_city}. For this, the Internet of Things (IoT), data analysis and machine learning technologies have been crucial \cite{smart_city2}. In addition to these, in recent literature, the Digital Twin (DT) technology has emerged as an enabler concept thanks to their real-time monitoring, comprehensive modelling, and cognitive attributes \cite{dt_transport_smart_grid}. 

Integrating DTs into smart cities is foreseen to drive the implementation of sustainable smart cities \cite{city_dt}. In these smart city models, real-time data and control flows are present between the physical environment and DT, which creates virtual dynamic replicas. In the creation of these replicas, DT  modelling challenges arise due to the dynamicity of the city environment. In DT modelling, a data-driven model of the physical environment needs to be formed to monitor, give context, and analyze in real-time. To achieve this, the continuous data flows from multiple sources should be processed and merged under one model while capturing an accurate snapshot of the environment in every update \cite{middleware}. In other words, the model is required to be in convergence with the actual environment. Here, ensuring this convergence stands as a great challenge, which stems from the updates to each entity in the model from the sources not corresponding to the same point in time.

Moreover, to ensure the DT can capture changes and be up-to-date, data has been gathered with higher frequencies \cite{t6conf}. However, this can result in high resource usage and energy consumption on DT due to frequent processing and model updates, which may be unnecessary if the received data does not reflect a change in the environment. Considering these, we identify the challenges as below.
\begin{itemize}
    \item \textbf{Accurate Modeling:} In smart city applications, especially in a dynamic transportation scenario, the conventional relation-based modelling is insufficient to serve exact twins \cite{duran2023digital}. In addition, asynchronous updates to these twins result in temporal misalignment within the model. Therefore, enhanced modelling methods should be elaborated to ensure the exact alignment of the twins with their physical counterparts.

    \item \textbf{Resource Intensive Twining:} The need for high-frequency data in smart cities may strain resources and cause increased energy consumption due to updates to the model that do not change the value. Therefore, adaptive data update methods should be developed for the right-time data that needs to be captured from such a dynamic environment with low energy consumption levels.
\end{itemize}


\section{Related Work}


DTs are real-time virtual copies of the physical environment with two-way real-time communication between the physical and virtual. They are utilized in many fields, including smart cities \cite{dtn_survey}. For example, in the study of \textit{Kusic} et al., the what-if analysis of smart city transportation systems is done to establish a regulation-based management system. This proposed system can work on a real-time data stream and provide timely control of speed limits \cite{KUSIC2023101858}. Moreover, there are efforts to realize these, as in the New South Wales Digital Twin project in which a city-wide DT is designed to manage emergency scenarios, natural resources, and environmental conditions \cite{nsw}.  In another study, DT modelling is performed in core networks to decrease the energy consumption of network discovery service to create green communication paradigms for smart cities \cite{tgcn}.

Despite the increasing studies on DTs, to the best of our knowledge, there has been limited research on the perspective of DT data modelling. \cite{dt_model} surveys the digital twin-oriented methodologies and points out that the efficiency in data processing and modelling in DT and convergence to the environment is crucial. Different approaches for data modelling in DTs have been utilized. For example,  \cite{smart_city_relational_databases} proposes a DT data framework for digital twins that uses relational databases. Moreover, in \cite{dtwn} and \cite{poc}, a semantic graph-encoded JavaScript Object Notation for Linked Data (JSON-LD) has been utilized. And some other information systems are also used, based on traditional data files \cite{gis_dt}. Despite the diversity in these approaches, the update methodologies used in these studies are to update the DT with the data upon arrival. However, this may lead to temporal misalignment between entities. In other words, different entities' data representing the same time point may be reflected with an offset. This may decrease the DT accuracy due to any given snapshot of the DT not aligning with the real environment.

Digital twins require high synchronization with the real environment, meaning the devices should send data more frequently in highly dynamic conditions. This may cause an increase in resource and energy consumption on the DT side. A similar tradeoff nature has been present between the Age of Information and energy in the literature of IoT \cite{aoi_energy}. However, in the case of DT, a tradeoff may occur between synchronization and energy efficiency of DT. Considering these, this paper aims to answer the research question \textit{“How to (i) ensure the accurate modelling of digital twin models and (ii) provide low-overhead and energy efficient updates considering smart city applications from a transportation perspective?”}. To answer this, we propose spatiotemporal graph modelling for transportation networks in green smart cities and reinforcement learning-based adaptive twining mechanisms. 

\subsection{Contributions}
We list the contributions as below:
\begin{itemize}
    \item We utilize spatiotemporal graphs that represent data in space and time domain, ensuring precise alignment of the model. Moreover, we implement this model using graph databases that is able to store nodes and edges in a graph-native approach.
    
    \item We introduce novel Reinforcement Learning-based Adaptive Twining (RL-AT) mechanisms that operate by the dynamic of the transportation networks. We utilize incoming, outgoing, and current traffic density features. By doing so, we can manage the streaming information flow and capture right-time data with lower overhead and energy consumption.
\end{itemize}

\section{Reinforcement Learning-based Digital Twin Model}

\begin{figure}
    \centering
    \includegraphics[width=0.85\linewidth]{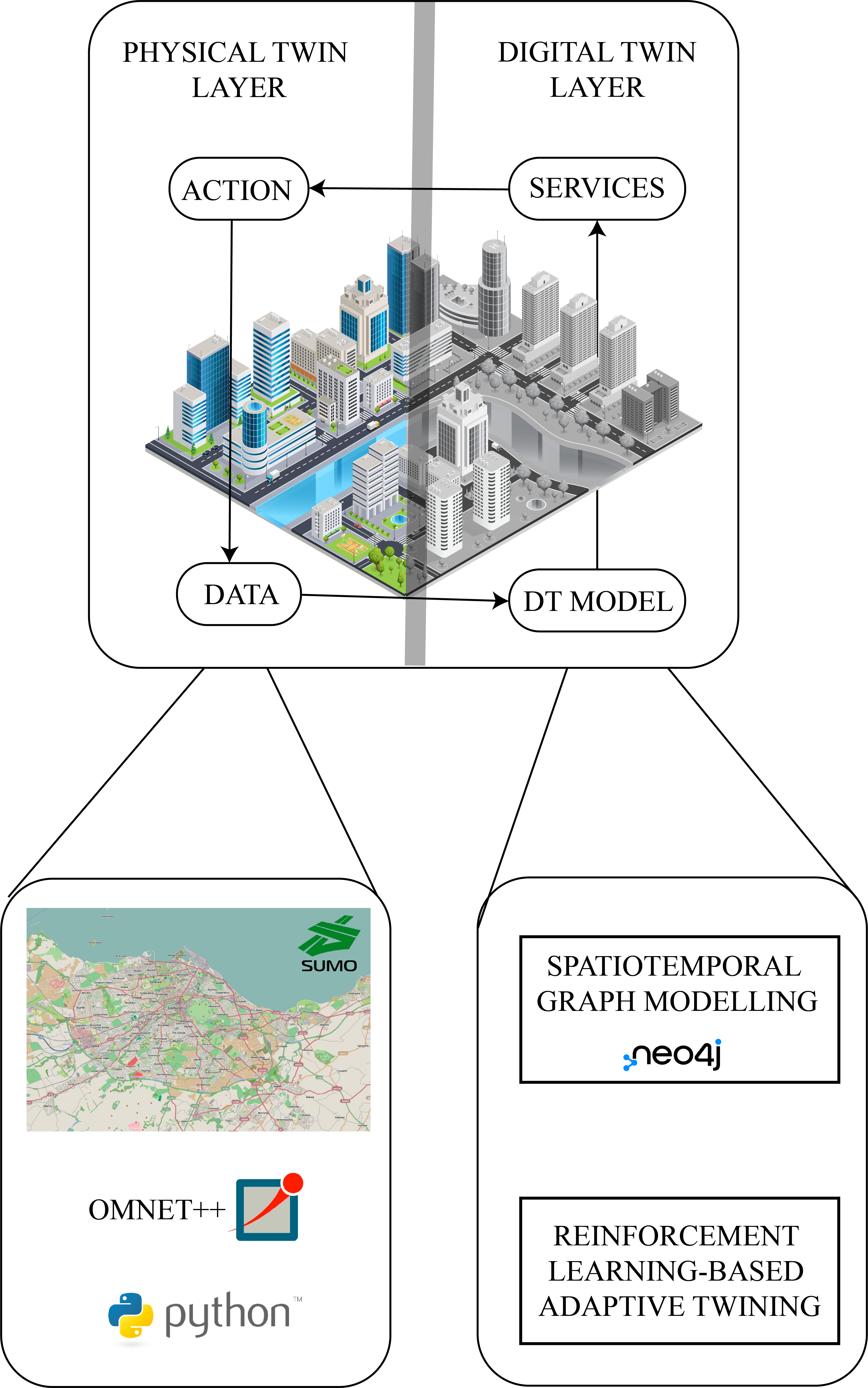}
    \caption{Reinforcement Learning-based Adaptive Digital Twin Model for Green Cities}
    \label{fig:System Architecture}
\end{figure}

The Reinforcement Learning-based Digital Twin Model consists of the Physical Twin Layer and the Digital Twin Layer, as shown in Figure \ref{fig:System Architecture}.

\subsection{Physical Twin Layer}

The Physical Twin Layer includes the city's infrastructural, technological, and natural elements. In the scope of this article, the transportation perspective is considered. Within this infrastructure, IoT devices, referred to as the Physical Twins (PTs), are deployed to gather data on the traffic flow to be sent to the DT. This study discusses the data-driven modelling and management of the updates to this model. Therefore, we assume this layer can provide any required information through a network infrastructure. We construct this layer using Simulation and Urban Mobility (SUMO)\cite{eclipse_sumo}. Moreover, we integrate the OMNET++ network simulator \cite{omnetpp} to simulate the underlying communication infrastructure.

\subsection{Digital Twin Layer}

Continuous data from the PTs are utilized to generate the real-time DT model. The data from distributed sources across the city ensures that the DT model can synchronize with the PTs. Moreover, services such as transportation management leveraging data analytics and machine learning techniques that can control the environment are present. 

In this, we propose the Spatiotemporal Graph Modelling and Reinforcement Learning-based Adaptive Twining (RL-AT) mechanism to achieve accurate modelling and handle the resource-intensive twining process.

\subsection{Spatiotemporal Graph Modelling}

The spatiotemporal graphs are created using the following building blocks.

\begin{definition}[Spatial Graph, $S$]
The weighted directed graph for spatial modelling is $S=(V, E,w)$ where $V$, $E$, and $w$ correspond to the vertex set, edge set, and weight function. 
\end{definition}

The physical infrastructure information is utilized to construct the spatial graph. Here, the graph weights represent the distances of each junction in the transportation network.

\begin{definition}[Graph Signal, $X$]
The feature matrix of the graph is represented as $X \in R^{NxFxT}$ where N is the number of vertices in the $S$ graph, F is the number of features, and T is the time steps. The graph signal in a given time $t$ is denoted as $X_t$.
\end{definition}
The graph signals are a spaced time series that includes the data collected from the PTs. Here, the values within any given $X_t$ matrix correspond to measurements of IoT devices in time $t$. Therefore, this ensures that any given snapshot of the model has each node in a uniform latency to the real environment. 

These spatiotemporal graph models denoted as $G$, are stored using the Neo4j graph database \cite{neo4j}. In Neo4j, each IoT device is represented as a node ($v \in V$) containing two types of relationships, namely, spatial relation and temporal relation and denoted as [:spatial] and [:temporal], respectively. Here, the collected data is stored in the node properties, the $w(e)$ in the property of [:spatial] relationship and the time difference in the property of [:temporal] relationship.

\subsection{Reinforcement Learning-based Adaptive Twining (RL-AT)}

\begin{figure}
    \centering
    \includegraphics[width=\linewidth]{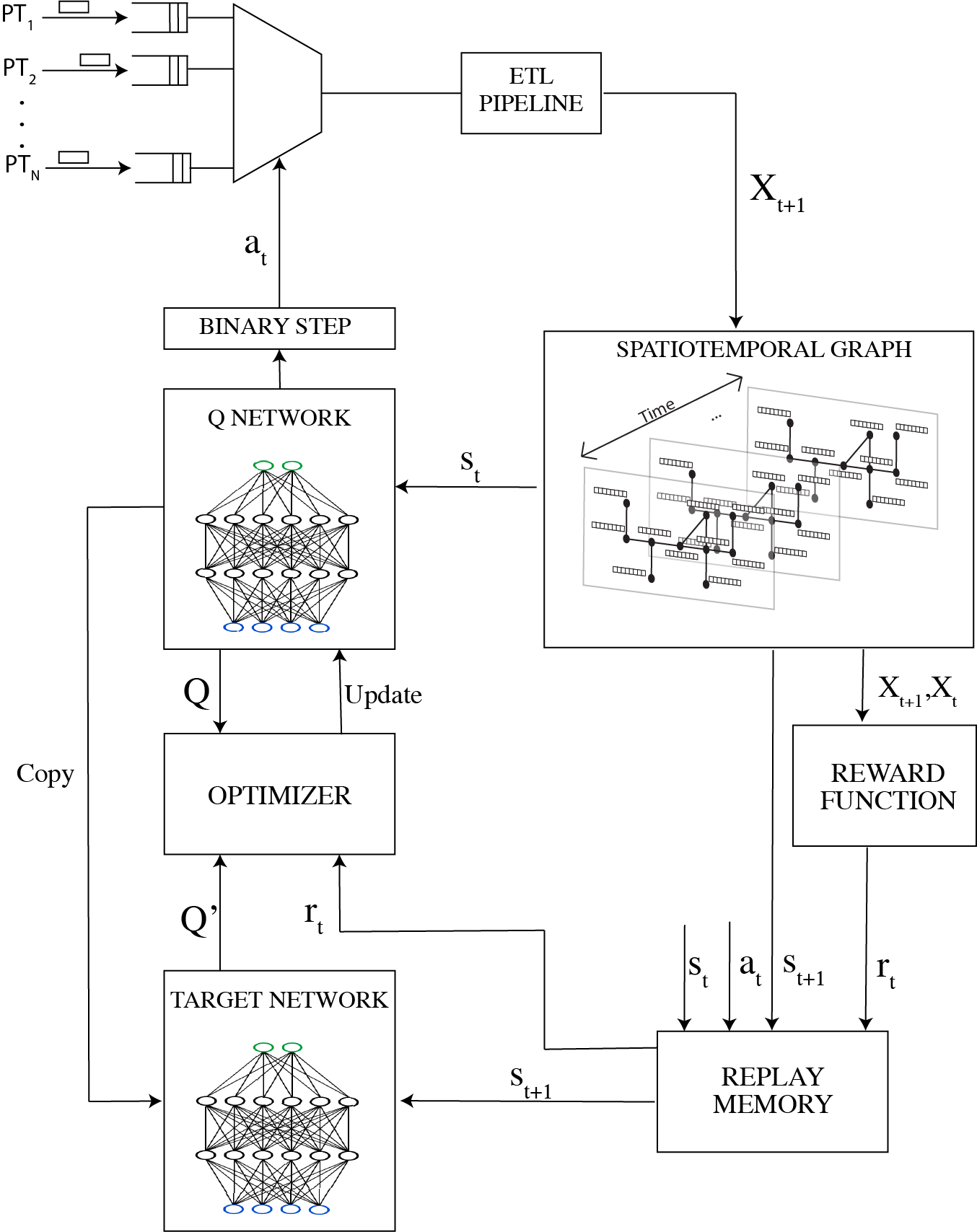}
    \caption{RL-AT Mechanism}
    \label{fig:rl-atr}
\end{figure}

The twinning rate is defined as the frequency of synchronization attempts of the physical object \cite{twining_rate}. This frequency may vary between devices due to external factors such as the information flow encountering different network latency and packet loss. Therefore, we define the RL-AT to ensure physical-to-virtual convergence in a transportation network while catching the dynamic flow variations energy-efficiently.

The RL-based Adaptive Twining dynamically adjusts the update strategy to achieve maximum DT accuracy while minimizing resource consumption. To do so, Deep Q Learning is used to decide which node to be updated. In this approach, we denote the state as $s_t$, which is equivalent to $(X_{ijk})_{ t-\delta < k \le t}$, the spatiotemporal graph signal for the $\delta$ time window. This time window is the number of past updates the agent considers when deciding the strategy.  Moreover, the $a_t$ is the update action that is a $N$ vector corresponding to whether an update will be done for the corresponding node and the actions space $A$ is. 
\begin{equation}
    A = \{a_{ij} | a_{ij} \in {0,1}\} 
\end{equation}

Furthermore, the reward ($r_t$) for the update action is calculated using the mean square error between the current values in DT ($X_t$) and the next $X_{t+1}$ with a penalty given based on the energy consumption associated with the action.

\begin{equation}
    r_t = \frac{1}{N+F} \sum_{i=0}^N\sum_{j=0}^F \sqrt{(X_{i,j,t+1}-X_{i,j,t})^2}    - P(a_t)
    \label{reward_function}
\end{equation}

Here, the $P(.)$ is the penalty function that considers memory operations needed to apply the action, and it is defined as

\begin{equation}
    P(a_t) = \sum_{i=0}^N a_t(1 + 1 + 1 + |\{e_{ij} | e_{ij} \in E\}|)
\end{equation}

This formulation considers the number of memory operations on data and terms in the equation corresponding to the retrieval and node and edge creations (Algorithm \ref{alg:rlat} line 8-24) in the spatiotemporal graph.

Figure \ref{fig:rl-atr} depicts the RL-based Adaptive Twining mechanism. The data flows coming from the PTs are queued before they are reflected in the model. Here, the queues from which the data will be updated or dropped are decided by the Q Network with an activation function as Binary Step. Then, the Extract Transform and Load (ETL) pipeline performs the model updates by incoming update data and creating the respective nodes, edges and properties in the Neo4j. Then, using Equation \ref{reward_function}, the reward associated with that action is calculated. The tuple, which represents the experience of the model $(s_{t}, a_t, r_t, s_{t+1})$, is stored in the replay memory, and these are then utilized in the optimization of the model.

This mechanism is performed according to the Algorithm \ref{alg:rlat}. The RL-AT algorithm retrieves the state information from the Neo4 to perform the decision-making in line 5 for the action. According to this, a node for the PT is created in line 11 with a temporal edge with the previously created node for that PT in line 13. Moreover, the nodes with a spatial edge with updated PT are also created, and edges are formed if they are to be updated in lines 10-13. If not, respective spatial edges are formed with the latest node for the spatially connected PT in lines 18-20. Here, the $Node_{t-kj}$ refers to the most recent created node for $j$ in the storage. Then, the reward associated with the action is calculated and used to optimise the DQN.

\begin{algorithm}[h]
  \caption{\textbf{RL-AT}: Reinforcement Learning-based Adaptive Twining}
  \label{alg:rlat}
  \begin{algorithmic}[1]
    \State Initialize $episodes$, $G$, $UpdateInterval$
    
    \For{e in $episodes$}
   
    \For{t in time}
    \State  Retrieve $s_t$
    \State Calculate $Q$ and $a_t$ using Q Network
    \State Apply $a_t$
    \State Perform ETL Pipeline to create $X_{t+1}$
    \For{$i$ in $N$}
    \If{$a_{t}[i]$ is 1}
        \If{$Node_{ti}$ is not present}
            \State Create $Node_{t+1i}$ using $X_{t+1}[i]$
        \EndIf
         \State Create  $Node_{t+1i}$-temporal- $Node_{ti}$  
        \For{$j$ in $N$}
            \If{$e_{ij}$ is 1 and $a_{t}[j]$ is 0}
                \State Create  $Node_{t+1i}$-spatial-$Node_{t-kj}$
            \Else
                \State Create $Node_{t+1j}$ using $X_{t+1}[j]$
                \State Create $Node_{t+1i}$-spatial-$Node_{t-kj}$ 
                \State Delete  $Node_{t+1i}$-spatial-$Node_{t-kj}$ 
            \EndIf
        \EndFor
        \EndIf
       \EndFor 
    \State $r_t \leftarrow$ Equation \ref{reward_function}
    \State Retrieve $s_{t+1}$
    \State Calculate $Q'$ for $s_{t+1}$ using Target Network
    \State Run Optimizer, Update Q Network

    \If{$t$ mod $UpdateInterval$ is $0$}
    \State Copy Q Network to Target Network
    \EndIf
   \EndFor
\EndFor
  \end{algorithmic}
\end{algorithm}

\section{Experimental Results}

\subsection{Simulation Setup}
The SUMO \cite{eclipse_sumo} and OMNET++ \cite{omnetpp} simulators are used to simulate the transportation network in green smart cities with the network infrastructure. In SUMO, the behaviour and road infrastructure are modelled with junctions serving as IoT devices corresponding to PTs. Here, the simulation scenario is generated using the data from OpenStreetMap \cite{osm}. This simulation is run with integration to OMNET++ simulation via the TraCI interface \cite{traci}. In OMNET++, the PTs are configured to send their data through the network to the DT device using the MQTT protocol. Via this simulation, the dataset of transportation behaviour and the packet delivery time datasets are generated. 

Moreover, a deterministic simulation environment is developed using Python to evaluate the proposed RL-based Adaptive Twining. Moreover, the proposed mechanism is implemented using the KerasRL library \cite{kerasrl}, and the spatiotemporal graph models are also implemented using Neo4j \cite{neo4j}.

\subsection{Performance Evaluation}
In this section, we aim to examine the performance of the Reinforcement Learning-based Digital Twin Model considering (i) DT accuracy performance of spatiotemporal graph and traditional modelling, (ii) querying performance of storage method of spatiotemporal graph model, (iii) cumulative reward of the RL-based Adaptive Twining under different learning rates, and (vi) average energy consumption and total RAM usage under changing payload sizes with and without RL-based Adaptive Twining.

In the first set of our experiments, we compare the accuracy of the spatiotemporal graph model to the traditional modelling mechanism that includes updating the entity directly. To investigate this, we compare the mean square error of the snapshots captured during the simulation with the snapshot of the physical under-changing values for $N$ in Table~\ref{tab:mse}. The results show a consistent error throughout, while traditional methods have an increasing error over the increasing number of nodes. This is because spatiotemporal graph modelling can preserve the time associated with the data within the model so that when a snapshot is taken, it is more convergent to the real conditions. On the other hand, the traditional method results in a temporal misalignment between the entities, causing higher MSE. The intensification over the increase of nodes may be the result of the data sent from the PTs to reach to the DT, making the misalignment worse.


\begin{table}[!h]
\centering
\caption{ MSE Values for Spatiotemporal Graph Modelling and Traditional Modelling}\label{tab:featureffect}
\begin{tabular}{|l|c|c|} 
\hline
N & \textbf{Spatiotemporal Graph} & \textbf{Traditional}  \\ 
\hline
20        & 2.1                 & 11 \\
40        & 2.4               & 16.4 \\
80        & 2.8                 & 19.1 \\
100        & 2.6                 & 20.8 \\ 
\hline
\end{tabular}
\label{tab:mse}
\end{table}

\begin{figure}[htbp]
\centerline{\includegraphics[width=\linewidth]{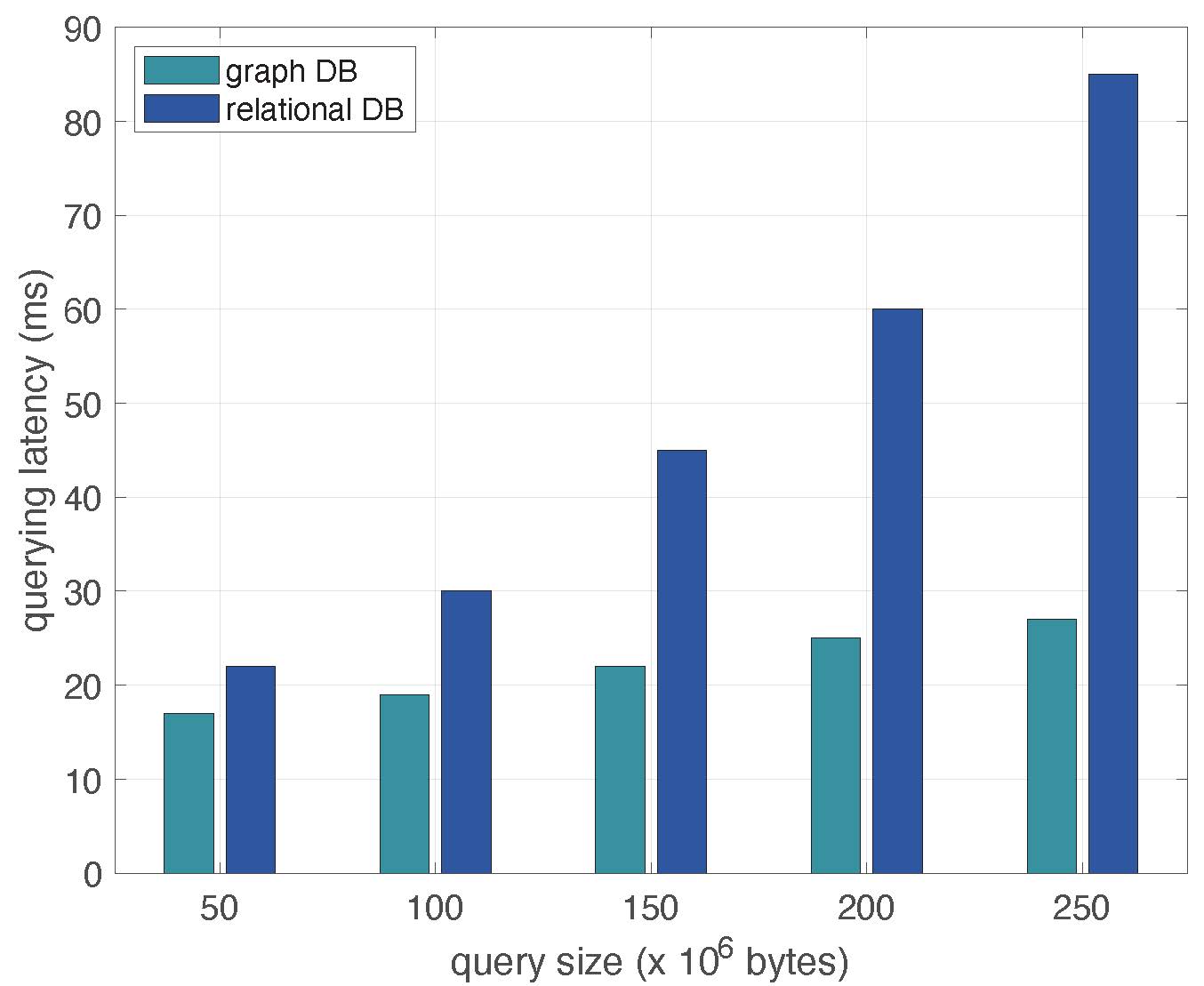}}
\caption{Querying latency for proposed spatio-temporal graphs and relational databases.}
\label{fig:qurying_latency}
\end{figure}
Then we evaluate the storage methods, graph database and relational database, for spatiotemporal graph model under increasing query size. As seen in Figure \ref{fig:qurying_latency}, when the query size increases, the relational database takes more time to get all the requested objects because of intensive join operations done on the data files. On the other hand, when a graph database is used, the expensive join operations are avoided by holding the relationships between nodes separately. According to the simulation results, we note 55\% decrease in the total querying latency with the usage of spatiotemporal graphs. This querying latency performance of graph databases can directly contribute to overhead reduction when generating the states. 

\begin{figure}[htbp]
\centerline{\includegraphics[width=\linewidth]{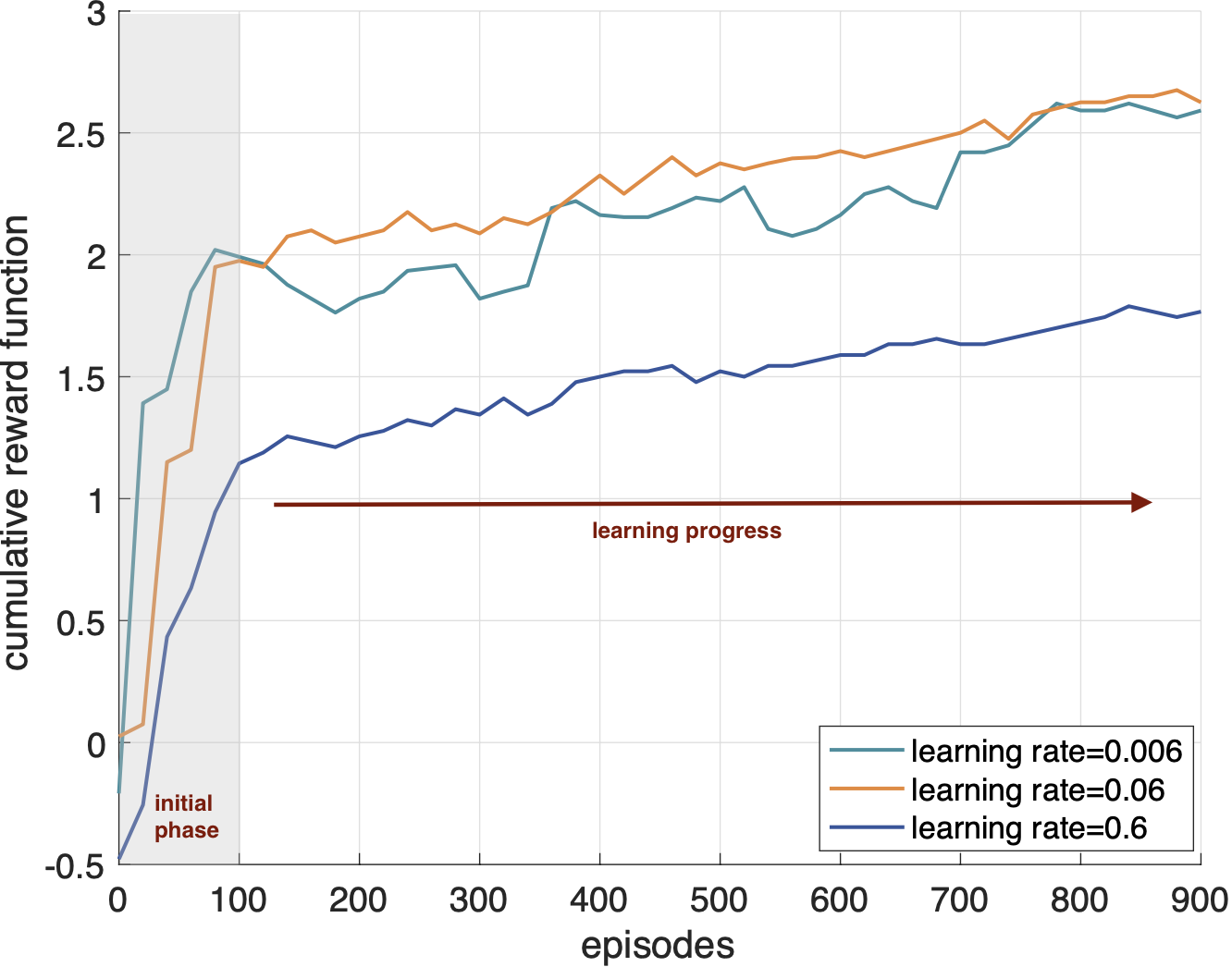}}
\caption{Cumulative reward function for DQN training episodes}
\label{fig:cumilative_reward}
\end{figure}

Moreover, we evaluate the cumulative reward of the RL-based Adaptive Twining under different learning rates in Figure \ref{fig:cumilative_reward}. Here, we observe that when the learning rate is equal to $0.6$, the algorithm cannot converge to a particular value, but convergence is achieved as it is decreased to $0.06$. This is because the lower learning rates can achieve improved generalization to unseen scenarios, which is prominent in the case of this paper due to the state space being extremely large.

\begin{figure}[htbp]
\centerline{\includegraphics[width=\linewidth]{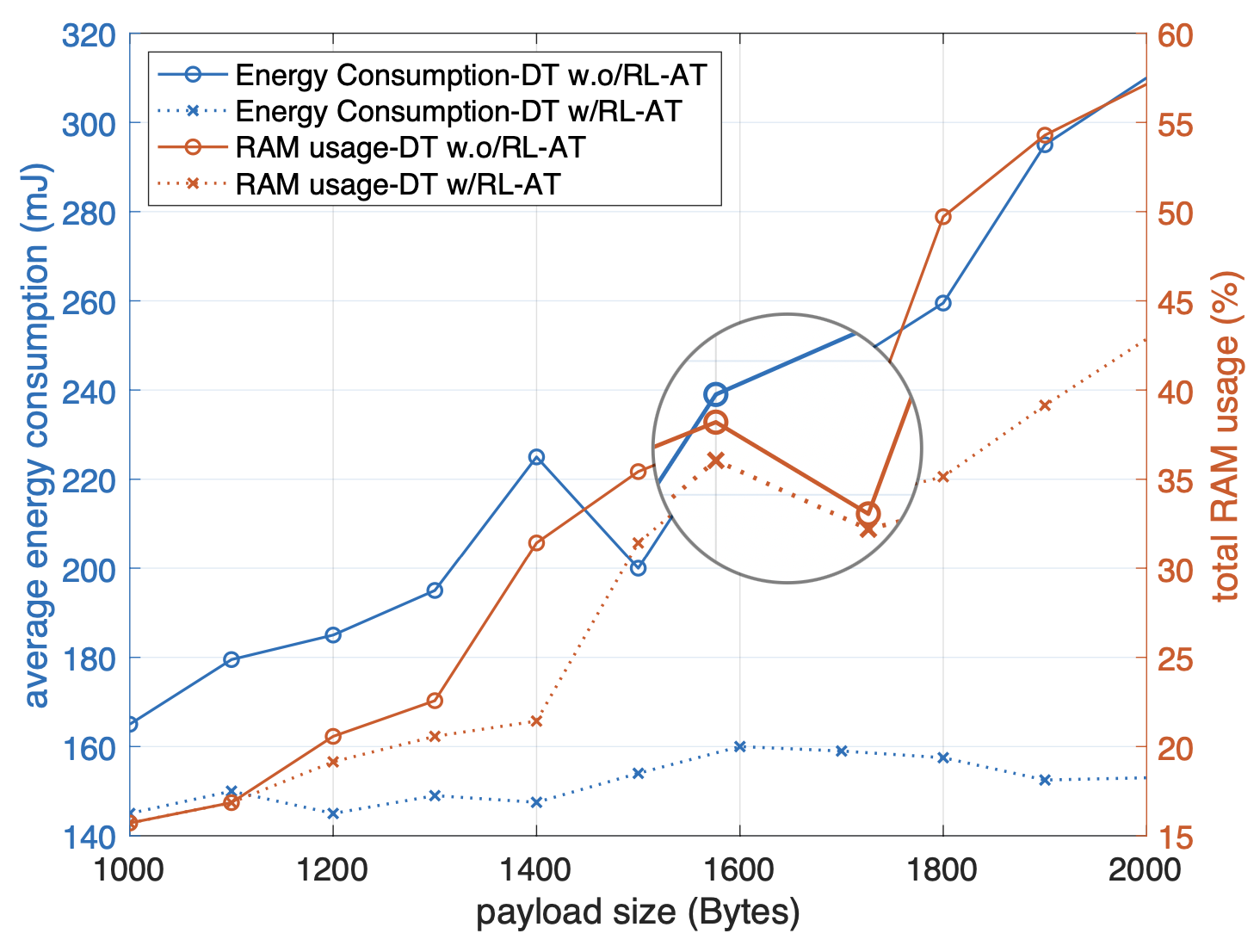}}
\caption{Average energy consumption (mJ) and RAM usage percentage comparison under changing payload sizes with and without RL-based Adaptive Twining}
\label{fig:energy}
\end{figure}

Furthermore, we observe DT's average energy consumption and total RAM utilization with and without  RL-based Adaptive Twining under changing payload sizes in Figure \ref{fig:energy}. The proposed mechanism selectively avoids unnecessary updates that will not significantly influence the accuracy of DT. Besides, as RL-based Adaptive Twining decides if an update will be performed based on the influence this update will have on the accuracy and resources, it can capture right-time data with 20\% lower overhead in terms of total RAM usage compared to the case where all asynchronous updates are performed. Even though the noted RAM usage values are very close to each other when the payload size is around $\sim 1600-1800$ bytes (as shown in the circle in Figure \ref{fig:energy}), the total average energy consumption value is decreased by 20\% with RL-based Adaptive Twining. Additionally, the results show a more prominent difference in energy consumption, which can be attributed to higher energy costs of memory operations and reward function in Equation \ref{reward_function} applying penalty for memory operations.

\begin{figure}[htbp]
\centerline{\includegraphics[width=\linewidth]{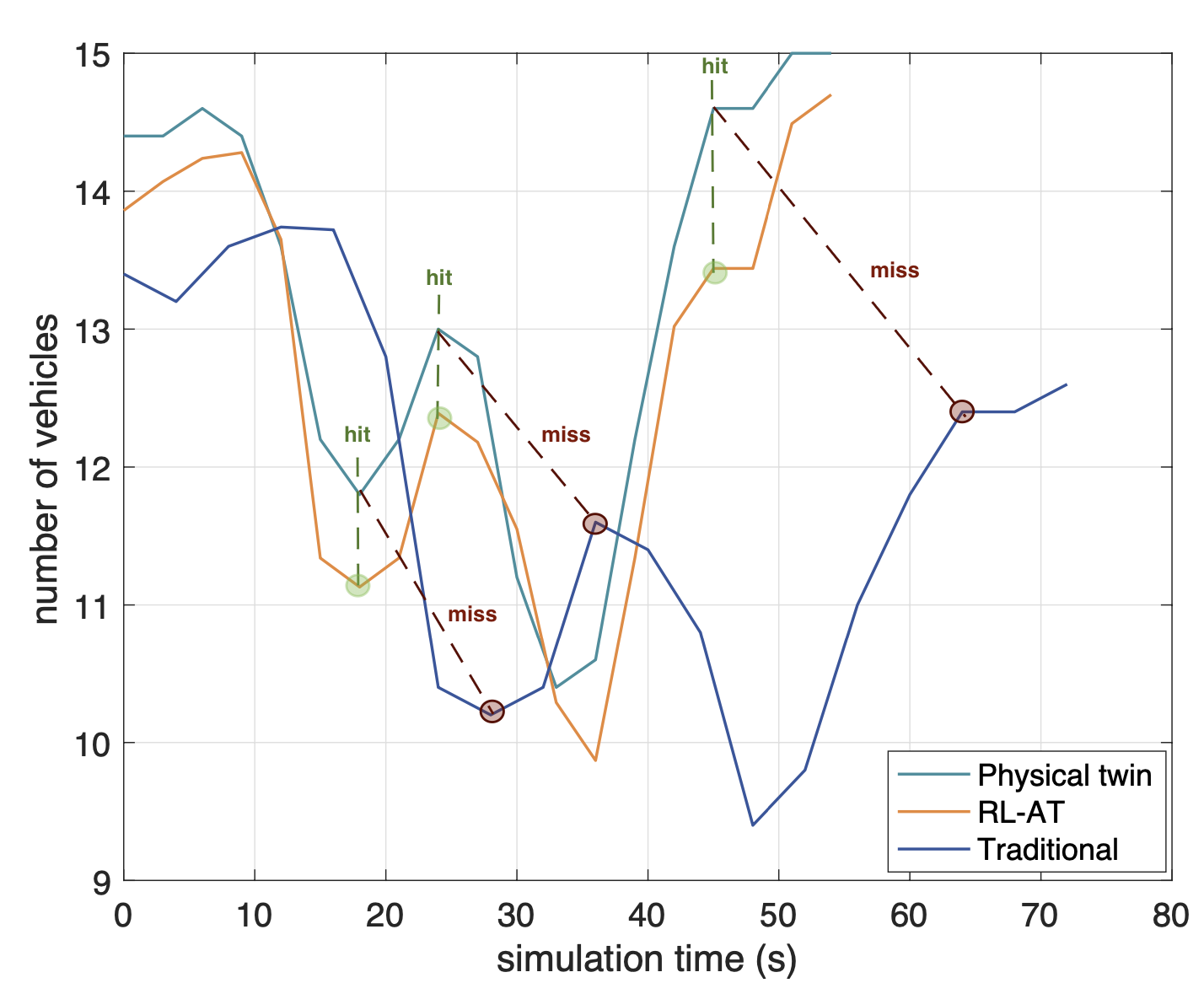}}
\caption{Synchronisation performance of the traditional method and RL-AT.}
\label{fig:syncperf}
\end{figure}

Lastly, we evaluate the synchronization by using the number of vehicles in the simulation as a baseline and measure the summation of differences in the incoming and outgoing traffic features as shown in Figure \ref{fig:syncperf}. Here, we sample three observations of physical twins and label them as \textit{hit} and \textit{miss} cases based on the time difference in which specific observations were captured on the DT side. The results show that both traditional and RL-AT follow a similar trajectory to baseline with a prominent shift of traditional that also increases during the runtime. This may be caused by the accumulation of data waiting for updates due to a bottleneck created by the available resources. This is avoided in RL-AT by using DQN to selectively choose which data from PT will be updated. Additionally, the difference of measurements in the \textit{hit} cases occurs because not every received data is reflected in the model. This can be considered a tradeoff between accuracy and timeliness. Here, the expectation from the traditional method was to achieve a lower difference of values when the trajectory shift is neglected. However, the observations show the contrary, possibly due to the temporal misalignment between entities in the model, as discussed previously in Table \ref{tab:mse}.





\section{Conclusion}

This study undertakes the challenge of accurate modelling and resource-intensive operations in green city digital twins. Firstly, utilize spatiotemporal graphs that mitigate the temporal misalignment. We implement this using graph databases, which can achieve 55\% higher querying performance than relational databases. Then, we propose the Reinforcement Learning-based Adaptive Twining (RL-AT) mechanism that utilizes Deep Q Networks (DQN). Here, we incorporated the penalty function defined based on memory operations into the reward function. Using this, we can capture right-time data, increasing the DT accuracy and 20\% lower RAM utilization and 25\% energy consumption. 

For future work, we aim to explore combining knowledge-driven approaches and multi-agent reinforcement learning that can enhance the decision-making process for the twining.

\section*{Acknowledgment}
This work was supported by DeepMind Scholarship Program and The Scientific and Technological Research Council of Turkey (TUBITAK) 1515 Frontier R\&D Laboratories Support Program for BTS Advanced AI Hub: BTS Autonomous Networks and Data Innovation Lab. Project 5239903.


\bibliographystyle{IEEEtran}
\bibliography{bibfile.bib}

\end{document}